\documentclass{article}

\usepackage[final]{nips}

\usepackage[utf8]{inputenc} % allow utf-8 input
\usepackage[T1]{fontenc}    % use 8-bit T1 fonts
\usepackage{hyperref}       % hyperlinks
\usepackage{url}            % simple URL typesetting
\usepackage{booktabs}       % professional-quality tables
\usepackage{amsfonts}       % blackboard math symbols
\usepackage{nicefrac}       % compact symbols for 1/2, etc.
\usepackage{microtype}      % microtypography
\usepackage{amsmath}
\usepackage{amstext}
\usepackage{amssymb}
\usepackage{graphicx}

\usepackage{color}
\usepackage{comment}

\usepackage{lineno,hyperref}
\usepackage{graphicx}
\usepackage{courier}  %Required
\usepackage{url}  %Required
\usepackage{graphicx}  %Required
\usepackage{amsfonts}
\usepackage{amsmath}
\usepackage{amssymb}
\usepackage{multirow}
\usepackage{makecell}
\usepackage{algorithm, algorithmic}
\usepackage{color}
\usepackage{comment}

\title{ATM: Adversarial-neural Topic Model}

\author{
{\textrm Rui Wang}\\
Southeast University\\
Nanjing, China, 210096\\
rui\_wang@seu.edu.cn
\and
\textbf{Deyu Zhou\footnote{the superscript means the corresponding author.}}\\
Southeast University\\
Nanjing, China, 210096\\
d.zhou@seu.edu.cn 
\and
\textbf{Yulan He}\\
University of Warwick\\
Coventry CV4 7AL, UK\\
yulan.he@warwick.ac.uk
}

\begin{document}
\maketitle

\begin{abstract}
Topic models are widely used for thematic structure discovery in text. But traditional topic models often require dedicated inference procedures for specific tasks at hand.  Also, they are not designed to generate word-level semantic representations. {\color{black}To address the limitations, we propose a neural topic modeling approach based on the Generative Adversarial Nets (GANs), called Adversarial-neural Topic Model (ATM) in this paper.  To our best knowledge, this work is the first attempt to use adversarial training for topic modeling}. The proposed ATM models topics with dirichlet prior and employs a generator network to capture the semantic patterns among latent topics. Meanwhile, the generator could also produce word-level semantic representations. Besides, to illustrate the feasibility of porting ATM to tasks other than topic modeling, we apply ATM for open domain event extraction. {\color{black}To validate the effectiveness of the proposed ATM, two topic modeling benchmark corpora and an event dataset are employed in the experiments. Our experimental results on benchmark corpora show that ATM generates more coherence topics (considering five topic coherence measures), outperforming a number of competitive baselines. Moreover, the experiments on event dataset also validate that the proposed approach is able to extract meaningful events from news articles}. 
\end{abstract}

\section{Introduction}

Topic models \cite{blei2012probabilistic,DBLP:journals/ipm/Chen17} underpin many successful applications within the field of Natural Language Processing (NLP). Variants of topic models have been proposed for different tasks including content analysis of e-petitions\cite{DBLP:journals/ipm/Hagen18}, topic-associated sentiment analysis \cite{lin2009joint}, event extraction from social media \cite{zhou2014simple,zhou2016jointly, zhou2017unsupervised} and product aspect mining \cite{DBLP:journals/ipm/XiaoJLZS18}. However, topic models typically rely on mean-field variational inference\cite{asuncion2009smoothing} or collapsed Gibbs sampling for model learning. A small change to the modeling assumption requires the re-derivation of the whole inference algorithm, which is mathematically arduous and time consuming. 

In recent years, word embeddings (such as Word2vec\cite{mikolov2013efficient}, GloVe\cite{pennington2014glove}, fastText\cite{joulin2016bag,bojanowski2016enriching} and probabilistic fastText\cite{athiwaratkun2018probabilistic}) have gained an increasing interest thanks to their improved efficiency in representing words as continuous vectors in a low-dimensional space. The resulting embeddings encode numerous semantic relations (similarity or analogies) and are helpful for NLP tasks\cite{DBLP:journals/ipm/Fernandez-Reyes18,DBLP:journals/ipm/HsuLCS18}. But the traditional topic models could not generate such word-level semantic representations. 

{\color{black}To overcome the limitation that traditional topic model often need sophisticated inference algorithm}, Neural Variational Document Model (NVDM) \cite{miao2016neural} was devised based on the Variational Auto-Encoder (VAE)\cite{kingma2013auto} and used a hidden layer to reconstruct the document by generating the words independently. However, the usage of gaussian prior over topics in NVDM may lead to incoherent and similar topics being generated. On the contrary, Srivastava \cite{srivastava2017autoencoding} proposed LDA-VAE, a neural topic model based on the VAE, in which the logistic normal distribution was employed as the prior over topics for topic generation. To further enhance the quality of the generated topic, Srivastava replaced the mixture assumption with a weighted product of experts at the word-level and proposed the ProdLDA. {\color{black} But both the LDA-VAE and the ProdLDA were not able to produce word-level semantic representations. Besides, the logistic normal prior used in LDA-VAE and ProdLDA also could not capture the multiplicity topical aspects in a document and result in generating bad topics. }

{\color{black}To overcome the limitations that the traditional topic models often need sophisticated inference algorithm and the exist neural based topic models could not generate coherent topic words.} In this paper, we propose the Adversarial-neural Topic Model (ATM) based on adversarial training. The principle idea is to use a generator network to learn the projection function between the document-topic distribution and the document-word distribution. Instead of providing an analytic approximation, as in traditional topic models, the ATM uses a discriminator network to recognize if the input document is real or fake and its output signal could help the generator to construct a more realistic document from a random noise drawn from a dirichlet distribution. Due to the flexibility of neural networks, the generator is capable of learning complicated non-linear distributions. And the supervision provided by the discriminator in the adversarial training phase will help the generator to capture the semantic patterns embedded in the latent topics. Besides, the connection weights between the embedding layer and the word distribution layer of the generator also encodes the semantic information and naturally provides distributed representations of words as side product. 

{\color{black} The objectives of our work in this paper are, more succinctly, as follows:}

{\color{black}
\begin{enumerate}
\item Traditional topic models based on gibbs sampling or variational inference often need sophisticated inference algorithms and obtain incoherent topics. We are interested in devising a novel neural-based topic model which could mine coherent topics from text corpora automatically in an unsupervised manner. To this end, based on the Generative Adversarial Net, we  propose the ATM model  which could extract the coherent topics among text corpus. 
\item From a practical perspective, we would like to devise a neural-based topic model which could be transplanted to other task easily with limited modification. For this purpose, we modify the topic generation process of the proposed ATM and employ it for open domain event extraction task \cite{zhou2015unsupervised}, experiments on news articles corpus shows that the proposed model is able to extract meaningful events and also verifies the portability of ATM.

\end{enumerate}
}

The practical significance of this work is that the proposed approach (ATM) could generate more coherent topics than the state-of-the-art
topic modeling approaches. Meanwhile, it could also produce semantic representations for each word in the vocabulary as side product, which is currently not supported by the compared models. Besides, the proposed ATM could be easily ported to other NLP task (such as open domain event extraction) with limited modification. The rest of the paper is organized as follows. Section 2 reviews the related literature on neural topic models and generative adversarial nets. In Section 3, we provide the details of the proposed Adversarial-neural Topic Model. Section 4 will introduce our evaluation corpora and our obtained experimental results. Finally, the paper is concluded in Section 5 with suggestions for further work.

\section{Related Work}

Our work is related to two lines of research, neural-based topic modeling and the Generative Adversarial Nets. Thus, we will next briefly introduce the related work in two domain separately. 

\subsection*{Neural-based Topic Modeling}

To overcome the difficult exact inference of topic models based on directed graph, Hinton \cite{hinton2009replicated} modified the Restricted Boltzmann Machines and proposed a replicated softmax model (called RSM). Inspired by the variational autoencoder, Miao \cite{miao2016neural} used the multivariate gaussian as the prior distribution of latent space and proposed the Neural Variational Document Model (NVDM) for text modeling. More recently, to deal with the inappropriate gaussian prior of topic distributions in NVDM, Srivastava \cite{srivastava2017autoencoding} proposed the LDA-VAE which approximated the dirichlet prior using a logistic normal distribution, and the usage of logistic normal prior could help to generate more coherent and diverse topics. Srivastava \cite{srivastava2017autoencoding} replaced the mixture assumption with a weighted product of experts at the word-level and proposed the ProdLDA which further improved topic coherence.  

\subsection*{Generative Adversarial Nets}

As a neural-based generative model, the Generative Adversarial Nets \cite{goodfellow2014generative} have been extensively researched from both theoretical and practical aspects. 

Theoretically, \cite{nowozin2016f} used the Fenchel conjugate to define the F-divergence and proposed the F-GAN to generalize its optimization objective. To precisely measure the distance between two high dimensional distributions, \cite{arjovsky2017wasserstein} defined the Earth Mover's Distance (Wasserstein distance) and gave a computational method based on the weight clipping mechanism. Along this line, \cite{gulrajani2017improved} improved the Wasserstein GAN by adding a gradient penalty loss and promoted the stability of adversarial training.  

In practical applications, {\color{black} GAN-based models have been extensively researched in computer vision community, especially in image generation scenario. To incorporate the conditional information, Mirza \cite{mirza2014conditional} employed the random noise together with label as input and proposed the Conditional-GAN to generate image under the {\color{black}supervision} of the annotated label. The deep convolutional neural network were employed as the generator and the discriminator in \cite{radford2015unsupervised} to improve the quality of generated image. And \cite{ledig2016photo} also used the GAN-basd approch to generate super-resolution image.} {\color{black}On the other hand,  many variants of GAN have been developed for NLP tasks. Such as text generation, a hot research area in NLP.} The sequence generative adversarial network (SeqGAN) proposed in \cite{yu2017seqgan} incorporated a policy gradient strategy to optimize the generation process. Based on the policy gradient, Lin \cite{lin2017adversarial} proposed the RankGAN to capture the rich structures of language by ranking and analysing a collection of human-written and machine-written sentences. To overcome the mode collapse when dealing with discrete data, Fedus \cite{fedus2018maskgan} proposed the MaskGAN which used an actor-critic conditional GAN to fill in missing text conditioned on the surrounding context.  Along this line, Wang \cite{wang2018sentigan} employed multiple generator network (each for one sentiment) and proposed the SentiGAN to generate texts of different sentiment labels. {\color{black}Hu \cite{hu2017toward} incorporated the VAE into GAN framework for text generation. Besides, \cite{miyato2016adversarial,li2018learning} improved the performance of semi-supervised text classification using adversarial training.  Zeng \cite{zeng2018adversarial} designed GAN-based models for distance supervision relation extraction. Wang \cite{wang-lee-2018-learning} incorporated the generative adversarial net into a encoder-decoder framework and proposed a GAN based model for text summarization. Yang \cite{NIPS2018_7959} employed the target domain language model into GAN framework to transfer style of text.}

{\color{black}Despite many successful applications using GAN-based approaches, none of these approaches tackles the topic modeling problem. We propose the first GAN-based topic model called ATM, which differs from the existing approaches to neural topic modeling in the following aspects: (1) Unlike the NVDM and the LDA-VAE which use either multivariate gaussian prior or logistic-normal prior for latent topics, ATM uses the dirichlet prior instead. It makes sure that ATM could provide K-dimensional noise and each capture certain semantic patterns in the text corpus; (2) Unlike most GAN-based text generation approaches, a generator network is employed by ATM to learn the projection function between the document-topic distribution and the document-word distribution, which essentially captures the semantic patterns among latent topics rather than generating text sequences; (3) Unlike the traditional topic model,  ATM is able to generate meaningful  word-level semantic representations as a side product.}

\begin{figure*}[!h]
\centering
\includegraphics[
  width=1\textwidth,
  keepaspectratio]
{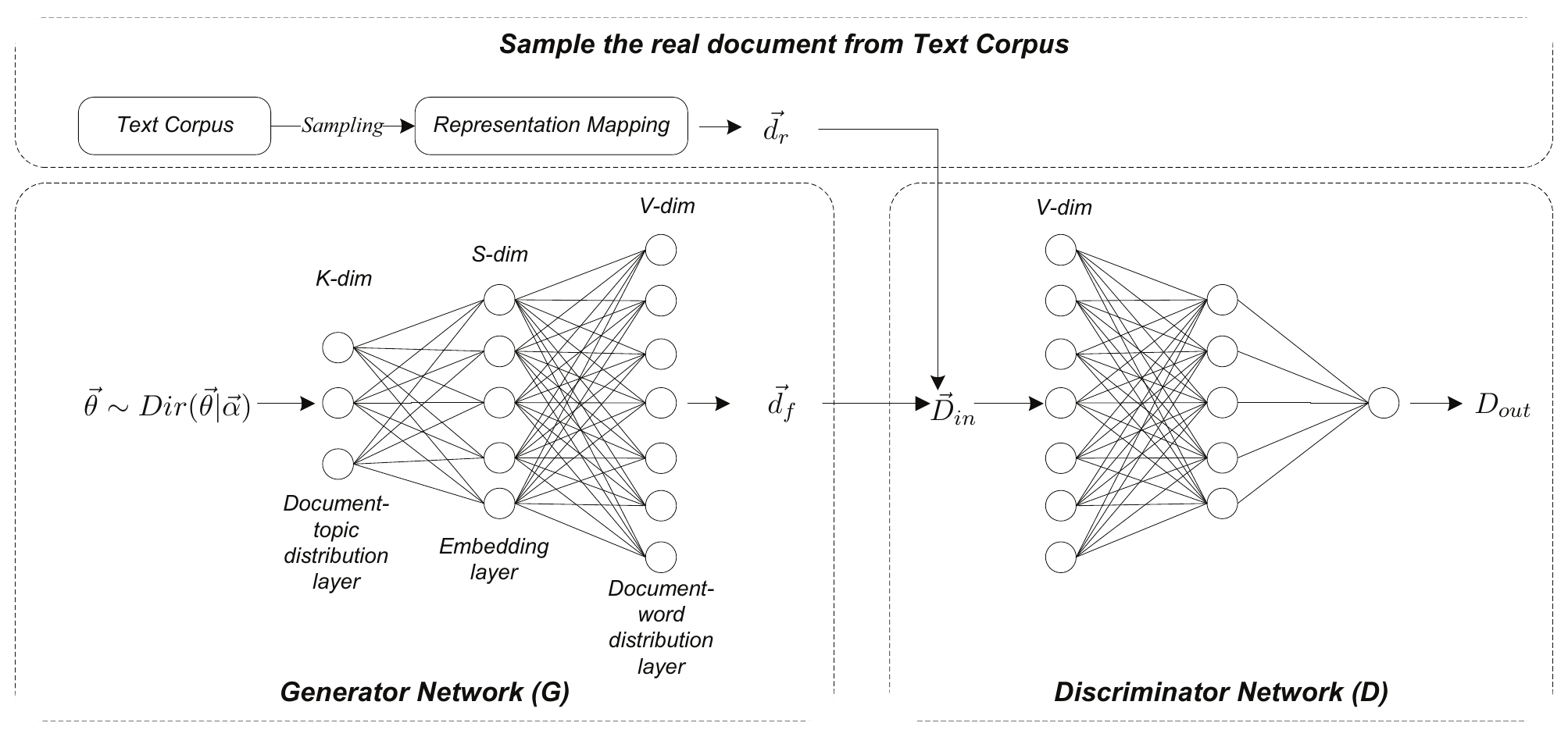}
\caption{The framework of the Adversarial-neural Topic Model (ATM).}
\label{fig:atm_framework}
\end{figure*}
\section{Adversarial-neural Topic Model}

We propose the Advesarial-neural Topic Model (ATM) as shown in Figure \ref{fig:atm_framework}. 
The proposed ATM contains three main components: (1) the document sampling module shown at the top of Figure \ref{fig:atm_framework}, which defines the representation mapping function and samples a real document $d_{r}\in\mathbb{R}^{V}$ from an input text corpus; (2) the generator $G$ takes a topic distribution $\vec\theta$ sampled from a dirichlet prior as input and generates the corresponding fake document $d_{f}$; (3) the discriminator $D$ takes $d_{f}$ and $d_{r}$ as input and discriminates the fake document from the real ones, whose output is subsequently used as a learning signal to update the parameters of $G$ and $D$. We explain the design and function of each of these modules in more details below.

\subsection{Representation Mapping}

Each document $d$ is represented by a normalized $V$-dimensional vector weighted by TF-IDF. More concretely:
\begin{align*}
tf_{i,d} & =\frac{n_{i,d}}{\sum_{v}n_{v,d}} \\
idf_{i} &=\log \frac{|C|}{|C_{i}|}\\
tf\textrm{-}idf_{i,d} &=tf_{i,d}\times idf_{i}\\ 
d_{r}^{i} &=\frac{tf\textrm{-}idf_{i,d}}{\sum_{v}tf\textrm{-}idf_{v,d}}
\end{align*}
where $V$ is the vocabulary size, $n_{i,d}$ denotes the number of times the $i$-th word appears in document $d$, $|C|$ denotes the total number of documents in the corpus, and $|C_{i}|$ is the number of documents containing the $i$-th word. With this representation, each document in the corpus could be  regarded as a multinomial distribution over $V$ words, and each dimension reflects the semantic coherence between the $i$-th word and the document $d$.

\subsection{Network Architecture}

The $G$ network contains three layers, the $K$-dimensional document-topic distribution layer, the $S$-dimensional embedding layer and the $V$-dimensional document-word distribution layer as shown in Figure \ref{fig:atm_framework}. First, the $G$ network takes a randomly sampled topic distribution $\vec\theta$ as input and transforms it into a document-word distribution. To model the multinomial property of the document-topic distribution, $\vec\theta$ is drawn from $Dir(\vec\theta|\vec\alpha)$:
\begin{align}
p(\vec\theta|\vec \alpha)= &Dir(\vec\theta|\vec\alpha)\triangleq\frac{1}{\triangle\left(\vec\alpha\right)}\prod_{k=1}^{K}\theta_{k}^{\alpha_{k}-1}
\end{align}
where  $\vec\alpha$ is the hyper-parameter of the dirichlet distribution, $\triangle(\vec\alpha)=\frac{\prod_{k=1}^{K}\Gamma(\alpha_{k})}{\Gamma(\sum_{k=1}^{K}\alpha_{k})}$, $K$ is the number of topics, $\theta_{k}\in[0,1]$ denotes the proportion of topic $k$ in the document and $\sum_{k=1}^{K}\theta_{k}=1$. 

Then, $G$ projects $\vec\theta$ into the $S$-dimensional (set to 100 in experiments) semantic space through the embedding layer based on equations :
\begin{align}
&\vec a_{s}=\max ((W_{s}\vec\theta+\vec b_{s}),leak*(W_{s}\vec\theta+\vec b_{s}))\\
&\quad\quad\quad\quad\quad \vec o_{s}=BN(\vec a_{s})
\end{align} 
where $W_{s}\in \mathbb{R}^{S\times K}$ is the weight matrix and $\vec b_{s}$ represents the bias term of the embedding layer, $\vec a_{s}$ is the state vector activated by the LeakyReLU function parameterized with $leak$, $BN$ denotes batch normalization and $\vec o_{s}$ is the output of the embedding layer. 

Finally, $G$ transforms $\vec o_{s}$ to a $V$-dimensional multinomial distribution $d_{f}$ using :
\begin{align}
&\vec h_{w}=W_{w}\vec o_{s}+\vec b_{w}\\
&o_{w}^{i}=\frac{\exp (h_{w}^{i})}{\sum_{v=1}^{V}\exp (h_{w}^{v})}
\end{align}
where $W_{w}\in \mathbb{R}^{V\times S}$ learns the semantic word embeddings and $\vec b_{w}$ represents the bias term, $\vec h_{w}$ is the state vector and $o_{w}^{i}$ denotes the probability of $i$-th word in $d_{f}$.

Likewise, we design the discriminator as a three layer fully connected network. The $D$ network employs the $d_{f}$ and the $d_{r}$ as input and outputs a scalar as shown in Figure \ref{fig:atm_framework}. A higher $D_{out}$ means that the discriminator is prone to consider the input data as a real document and vice versa.

\subsection{Training}

The fake document $d_{f}$ and the real document $d_{r}$ shown in Figure \ref{fig:atm_framework} could be viewed as the random sample from two $V$-dimensional dirichlet distribution $\mathbb{P}_{g}$ and $\mathbb{P}_{r}$. And the training objective of ATM is to let the generated distribution $\mathbb{P}_{g}$ approximate the real data distribution $\mathbb{P}_{r}$ as much as possible. Thus, the choice of divergence that measures the distance between two distributions is crucial for effective training of ATM. 

The original GAN \cite{goodfellow2014generative} used the Jensen-Shannon divergence as the optimization objective. However, \cite{arjovsky2017wasserstein} argued that the divergences which GANs typically minimize are potentially not continuous with respect to the generator's parameters, leading to mode collapse and training difficulty. They proposed instead using the Earth-Mover's distance (also called Wasserstein-1) which is defined as the minimum cost of transporting mass in order to transform the distribution $\mathbb{P}_{g}$ into the distribution $\mathbb{P}_{r}$. Further, \cite{gulrajani2017improved} improved the Wassertein-1 with a gradient penalty strategy which performed more stable. We follow their work and define the objective of ATM as:
\begin{align}
&L_{d} = \underset{d_{f}\sim \mathbb{P}_{g}}{\mathbb{E}}\left[D(d_{f})\right]-\underset{d_{r}\sim \mathbb{P}_{r}}{\mathbb{E}}[D(d_{r})]\\
&\quad L_{gp}= \underset{\hat{d}\sim\mathbb{P}_{\hat{d}}}{\mathbb{E}}[(\parallel\nabla_{\hat{d}}D(\hat{d})\parallel_2 -1)^{2}]\\
&\quad\quad\quad\quad L = L_{d}+\lambda L_{gp}
\end{align}
\begin{algorithm}[!h]
	\renewcommand{\algorithmicrequire}{\textbf{Input:}}
	\renewcommand{\algorithmicensure}{\textbf{Output:}}
	\caption{Training procedure for ATM}
	\label{alg:1}
	\begin{algorithmic}[1]
		\REQUIRE $K$, $\lambda$, $n_{d}$, $m$, $\alpha_{1}$, $\beta_{1}$, $\beta_{2}$
		\ENSURE the trained generator network $G$.
		\STATE Initial $D$ parameters $\omega_{d}$ and $G$ parameter $\omega_{g}$
		\WHILE{$\omega_{g}$ has not converged}
		\FOR{$t=1,...,n_{d}$}
		\FOR{$j=1,...,m$}
		\STATE Sample $d_{r}\sim \mathbb{P}_{r}$, 
		\STATE Sample a random  $\vec\theta\sim Dir(\vec\theta|\vec\alpha)$ 
		\STATE Sample a random number $\epsilon\sim U[0,1]$
		\STATE $d_{f}\leftarrow G(\vec\theta)$
		\STATE $\hat{d}\leftarrow \epsilon d_{r}+(1-\epsilon) d_{f}$
		\STATE {\color{black}$L_{d}^{(j)}=D(d_{f})-D(d_{r})$}
		\STATE $L_{gp}^{(j)}=(\parallel \nabla_{\hat{d}}D(\hat{d}) \parallel-1)^{2}$
		\STATE $L^{(j)}\leftarrow L_{d}^{(j)}+\lambda L_{gp}^{(j)}$
		\ENDFOR
		\STATE $\omega_{d}\leftarrow Adam(\nabla_{\omega_{d}}\frac{1}{m}\sum_{j=1}^{m}L^{(j)},\omega_{d},p_{a}) $
		\ENDFOR
		\STATE Sample $m$ noise $\left\{ \vec\theta^{(j)}\sim Dir(\vec\theta|\vec\alpha) \right\}$
		\STATE $\omega_{g}\leftarrow Adam(\nabla_{\omega_{g}}\frac{-1}{m}\sum_{j=1}^{m}D(G(\vec\theta^{(j)})),\omega_{g},p_{a})$
		\ENDWHILE 	
	\end{algorithmic}
\end{algorithm}
where $L_{d}$ and $L_{gp}$ denote the loss of discriminator $D$ and the gradient penalty, respectively, $\lambda$ is the gradient penalty coefficient, $\hat{d}$ could be obtained by sampling uniformly along a straight line between a real document $d_{r}$ and a generated document $d_{f}$, and $\mathbb{P}_{\hat{d}}$ is the distribution from which $\hat{d}$ is sampled.

{\color{black}In each training step, the same number of $d_r$ and $d_f$ samples are fed into the Discriminator and the distance between $\mathbb{P}_{g}$ and $\mathbb{P}_{r}$ is estimated using Eqs. 6-8. Thus, G and D networks could be updated to minimize the distance between $\mathbb{P}_{g}$ and $\mathbb{P}_{r}$.
 } Based on the model structure and the optimization objective described above, the training procedure for ATM is given in Algorithm 1. Here, $n_{d}$ denotes the number of discriminator iterations per generator iteration, $m$ represents the batch size, $\alpha_{1}$ is the learning rate, $\beta_{1}$ and $\beta_{2}$ are other hyper-parameters of Adam optimizer \cite{kingma2014adam}, and $p_{a}$ denotes $\left\{\alpha_{1},\beta_{1},\beta_{2}\right\}$. We use the default values of $\lambda=10$, $n_{d}=5$, $m=512$. Moreover, the $\alpha_{1}$, $\beta_{1}$ and $\beta_{2}$ are set to 0.0001, 0 and 0.9 respectively.

\subsection{Topic Generation}

The trained generator $G$ learns the projection function between the document-topic distribution and the document-word distribution. That is, given a topic distribution $\vec\theta_{d}$ for a document $d$, $G$ is able to generate the corresponding word distribution.  

To generate the word distribution of each topic, we use $\vec ts_{(k)}$, a $K$-dimensional vector, as the one-hot encoding of the $k$-th topic. For example, $\vec ts_{(1)} = [1, 0, 0, 0, 0]^{\intercal}$ in the five topic number setting. We could then obtain the word distribution $\vec \phi_{k}$ for topic $k$ using:
\begin{equation}
\vec \phi_{k}=G(\vec ts_{(k)})
\end{equation}

\section{Experiments}

We evaluate our proposed ATM on two tasks, topic extraction and open domain event extraction. We first describe the datasets and the baseline approaches, and then present the topic coherence evaluation results for the topic extraction task. Finally, we discuss the results of using ATM for open domain event extraction to validate the feasibility of applying ATM for tasks other than topic modeling.

\subsection{Experimental Setup}

Two publicly accessible datasets, Grolier\footnote{https://cs.nyu.edu/$\sim$roweis/data/ \label{grolier}} and NYtimes\footnote{http://archive.ics.uci.edu/ml/datasets/Bag+of+Words \label{nytimes}} datasets, are used for topic coherence evaluation, and an event dataset built based on the Global Database of Events, Language, and Tone (GDELT)\footnote{http://data.gdeltproject.org/events/index.html \label{event}} is used for event extraction. Details are summarized below: 
\begin{itemize}
{\color{black}
\item \emph{Grolier dataset}\textsuperscript{\ref{grolier}} is built from Grolier Multimedia Encyclopedia, and its content covers almost all the fields in the world, such as sports, economics, politics and etc. It contains 29,762 documents and is a benchmark text corpora in topic modeling. 

\item \emph{NYtimes dataset}\textsuperscript{\ref{nytimes}} is a collection of newswire articles written and  published by New York Times between January 1, 1987 and June 19, 2007 with article metadata provided by the New York Times Newsroom. This corpus also has a wide range of topics in real world, such as politics and entertainment.

%,  and it includes 99,992 documents.

\item \emph{Event dataset}.  This dataset is the subset of GDELT which is released by Google. we crawl the Database\textsuperscript{\ref{event}} and built the event dataset by selecting the articles published on the first day of May in 2014. It contains many real events occurred at that day, such as MH370 and Indian Election. 

%We crawl and parse the GDELT Event Database\textsuperscript{\ref{event}} containing articles published on the first day of May in 2014. 
}
\end{itemize}
\begin{table}[h]
\centering
%\small
%\scriptsize
\caption{The statistics of datasets.}
\label{tbs:statistic}
\begin{tabular}{lrr}
\hline
{\bfseries Dataset}& {\bfseries \#Document}&{\bfseries \#Words} \\
\hline
Grolier& 29,762& 15,276\\
NYtimes  & 99,992&12,604\\
Event&20,199& 9,346\\
\hline
\end{tabular}

\end{table}

We choose the following five models as the baselines: 
\begin{itemize}
{\color{black}
\item \textbf{LDA}~\cite{blei2003latent}, is a topic model that generates topics based on word the co-occurrence patterns from documents. With the usage of dirichlet prior topic distribution and word distribution, LDA could capture the multiplicity topic aspects from document collections in an unsupervised manner. We implement the LDA model and set the dirichlet prior of the document-topic distribution $\alpha=50/K$ and the dirichlet prior of the topic-word distributions $\beta=0.01$, following what have been suggested in \cite{griffiths2004finding}.
\item \textbf{NVDM}~\cite{miao2016neural}, is an neural based approach which models topics using variational auto-encoder. In NVDM, multivariate gaussian distribution is used as prior distribution of the latent space, and it is trained under the supervision of evidence lower bound (ELBO). We use the original implementation\footnote{https://github.com/ysmiao/nvdm}.
\item \textbf{LDA-VAE}~\cite{srivastava2017autoencoding}, is a neural topic model based on variational auto-encoder. To obtain readable topics, LDA-VAE substitute multivariate gaussian with  a logistic normal distribution as the prior of the latent space. In this paper, the original implementation\footnote{https://github.com/akashgit/autoencoding\_vi\_for\_topic\_models\label{vae}} of LDA-VAE is employed to obtain the  compared results. 
\item \textbf{ProdLDA}~\cite{srivastava2017autoencoding}, is a variant of LDA-VAE which also uses logistic normal as the prior of the latent space.  Beside, it assumes that the distribution over individual words is a product of experts rather than the mixture model used in LDA. The original implementation is used in this paper. 
\item \textbf{LEM} \cite{zhou2014simple}, is a bayesian modeling approach for open domain event extraction. It treats an event as a latent variable and models the generation of an event as a joint distribution of its individual event elements (organization , location , person , keyword)\footnote{means organization, location, person and keywords. \label{olpk}}. We implement the algorithm with the default configuration.
}
\end{itemize}

For the NYtimes dataset, we random select 100,000 articles and remove the low frequent words. For the Event dataset, we use the Stanford Named Entity Recognizer\footnote{https://nlp.stanford.edu/software/CRF-NER.html\label{ner}} \cite{finkel2005incorporating} for identifying the named entities (Location, Organization and Person). In addition, we remove common stopwords and only keep the recognized name entities and the tokens which are verbs, nouns, or adjectives from these event documents. The statistics of the processed corpora are shown in Table \ref{tbs:statistic}.

\begin{table*}[!ht]
\centering
%\scriptsize
\small
\label{table:stu}
\caption{Average topic coherence on Grolier and NYtimes corpus with five topic settings [20, 30, 50, 75, 100].}
\label{tbs:average}
\begin{tabular}{l|l|cccccc}
\hline
{\bfseries Dataset}&{\bfseries Model}& {\bfseries C\_P}&{\bfseries C\_A}&{\bfseries NPMI}&\bfseries UCI& \bfseries UMass \\
\hline
\multirow{5}{*}{Grolier}&NVDM&-0.187746 &0.145684 &-0.061911 &-2.114927 &-4.291624\\
&LDA-VAE&-0.220548 &0.150469  &-0.065378 &-2.479750 &-4.755522\\
&ProdLDA&-0.037436 &0.173391 &-0.019347 &-1.639878 &-4.542689\\
&LDA&0.190845 &0.200942  &0.049753 &-0.050336 &-2.918612\\
&ATM&\textbf{0.210448} &\textbf{0.218898} &\textbf{0.058167} &\textbf{0.105086} &\textbf{-2.765081}\\
\hline
\multirow{5}{*}{NYtimes}&NVDM&-0.413086 &0.134154  &-0.143711 &-4.307269 &-5.931614\\
&LDA-VAE&-0.157560 &0.148221  &-0.061418 &-2.420816 &-4.640276\\
&ProdLDA&-0.003455 &0.196395  &-0.028223 &-1.917367 &-4.193377\\
&LDA&0.308336 &0.212750 & 0.077278 &0.516503 &-2.420221\\
&ATM&\textbf{0.356771} &\textbf{0.237524}&\textbf{0.089874} &\textbf{0.658218} &\textbf{-2.324093}\\
\hline
\end{tabular}

\end{table*}

\begin{figure}[!ht]
\centering
\includegraphics[
  width=\textwidth,
  keepaspectratio]
{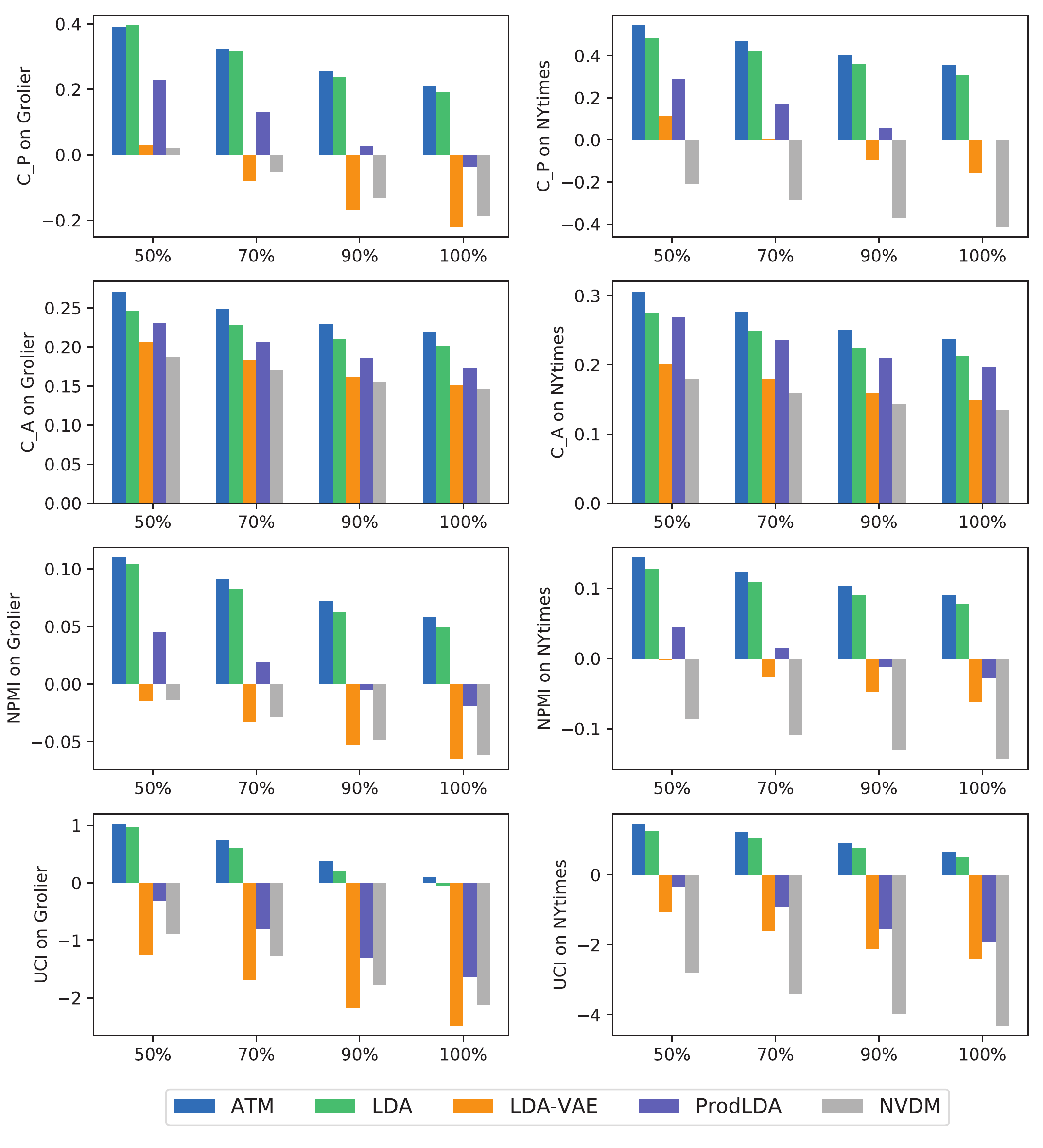}
\caption{Average topic coherence on Grolier and NYtimes with five topic settings [20, 30, 50, 75, 100] among topics whose coherence values are ranked at the top 50\%, 70\%, 90\% and 100\% positions.}
\label{fig:avg_bar}
\end{figure}

\subsection{Topic Coherence Evaluation}

\begin{figure}[h!]
\centering
\includegraphics[
  width=\textwidth,
  keepaspectratio]
{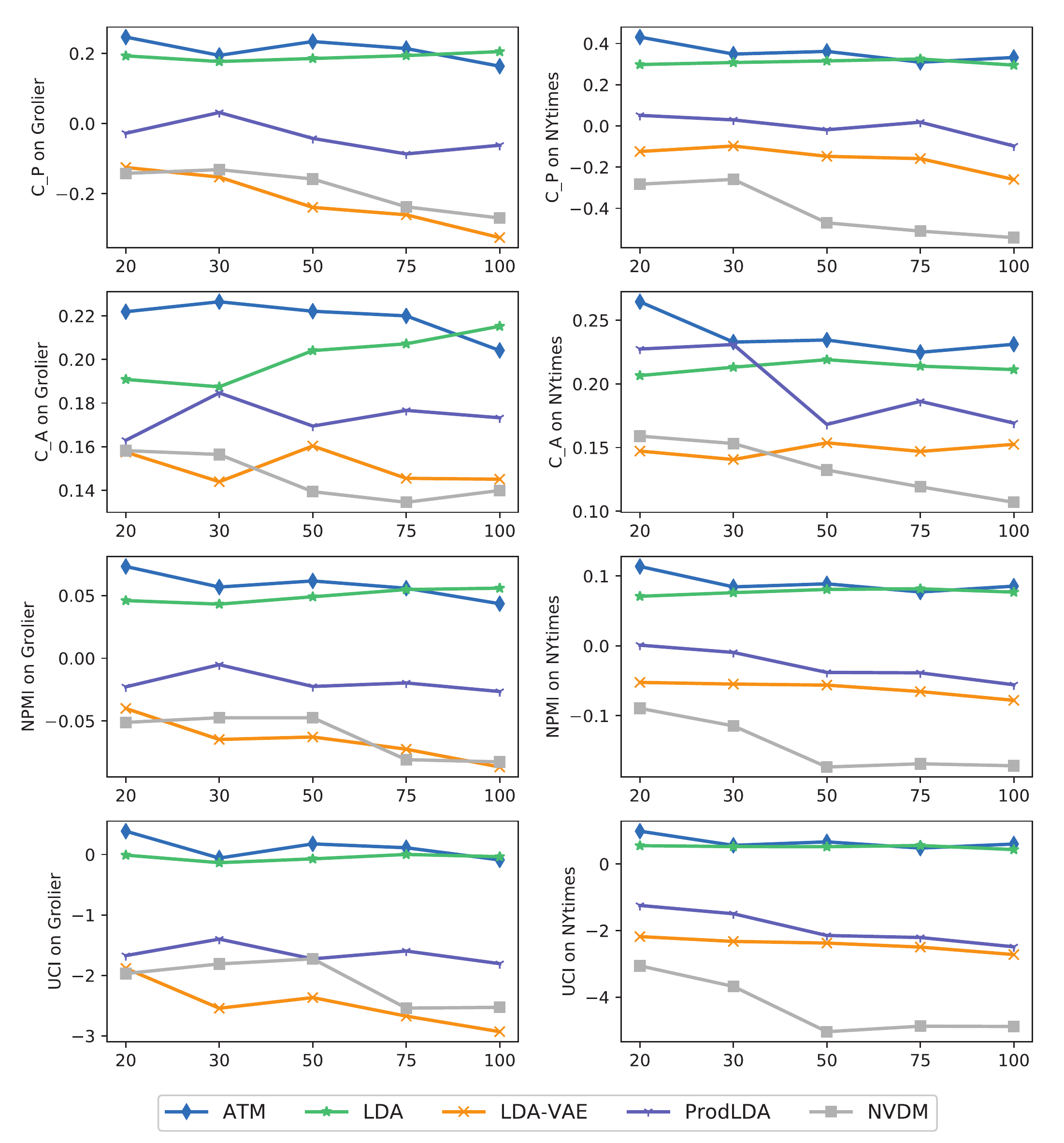}
\caption{Average topic coherence (100\%) on Grolier and NYtimes datasets vs. different topic setting [20, 30, 50, 75, 100].}
\label{fig:avg_curve}
\end{figure}

\begin{figure}[!b]
\centering
\includegraphics[
  width=\textwidth,
  keepaspectratio]
{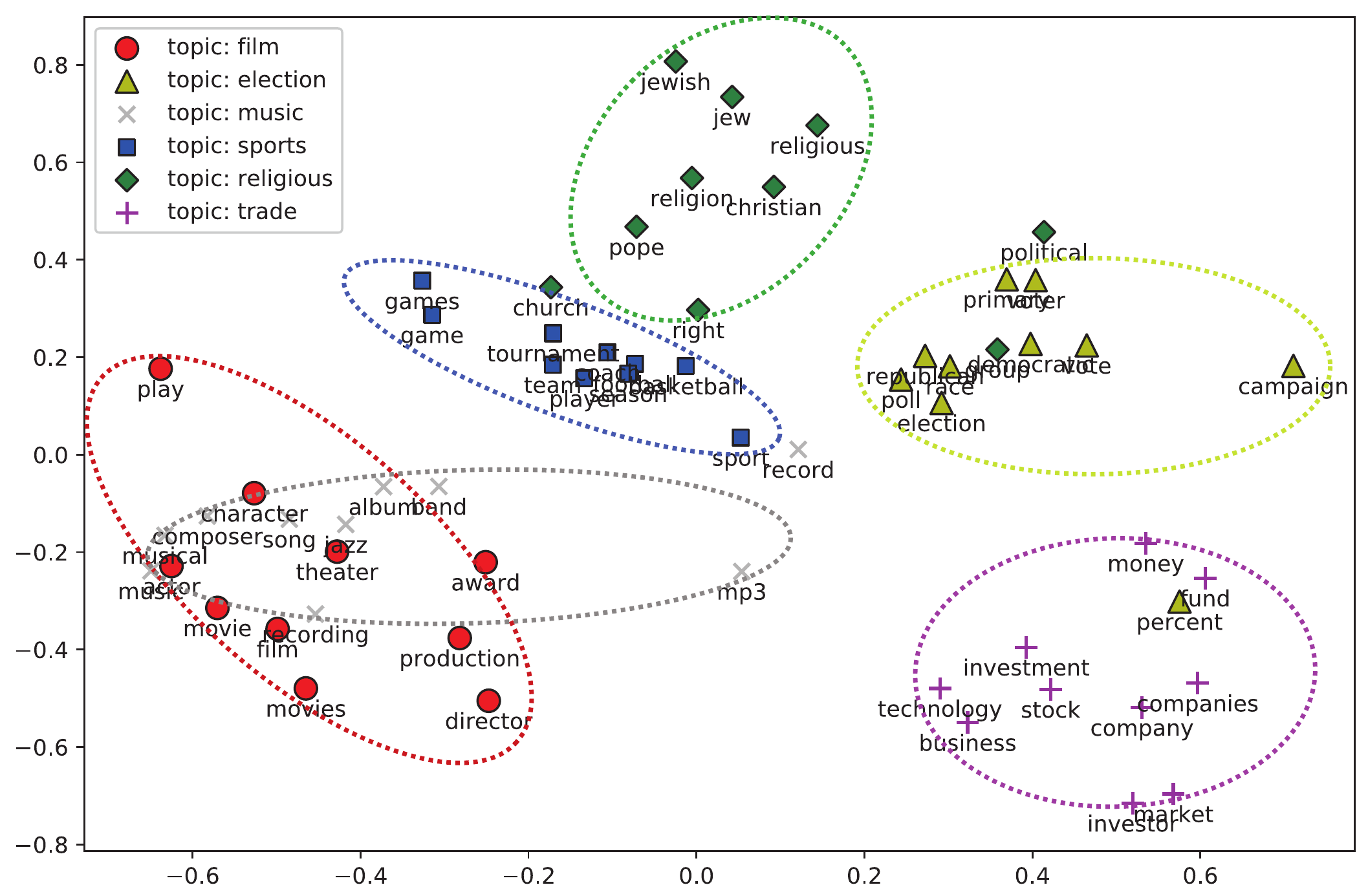}
\caption{visualization of the topic words from the six selected topics.}
\label{fig:topic_visualization}
\end{figure}

\begin{table*}[h]
\centering
\footnotesize
%\scriptsize
\label{table:stu}
\caption{Topic examples of all the models, italics means out-of-topic.}
\label{tbs:example_topics}

\scalebox{0.84}{
\begin{tabular}{c|l}
\hline
{\bfseries Model}& \multicolumn{1}{c}{\bfseries Topics}\\
\hline
\multirow{5}{*}{ATM}&\textbf{jet flight airline hour plane passenger trip plan travel pilot}  \\
& \textbf{stock market companies money investor technology fund investment company business}\\
& \textbf{music song musical album jazz band record recording mp3 composer}\\
& \textbf{voter vote poll republican race primary percent election campaign democratic}\\
& \textbf{film movie actor director award movies character theater production play}\\
\hline
\multirow{5}{*}{LDA} & \textbf{flight plane} \emph{ship} \emph{crew} \textbf{air pilot hour} \emph{boat} \textbf{passenger airport}\\
& \textbf{stock market} \emph{percent} \textbf{investor analyst} \emph{quarter} \textbf{investment shares share fund}\\
& \textbf{music song band sound record artist album show musical rock}\\
& \textbf{voter vote poll election campaign primary candidates republican race party}\\
& \textbf{film movie character play actor director movies} \emph{minutes} \textbf{theater} \emph{cast}\\
\hline
\multirow{5}{*}{ProdLDA}& \emph{wireless} \textbf{customer} \emph{telecommunication} \textbf{airlines} \emph{broadband satellites phones subscriber} \textbf{airline} \emph{provider}\\
& \textbf{brokerage securities broker lender buyer transaction investor investment stock borrower}\\
& \textbf{musical album} \emph{playwright} \textbf{composer} \emph{choreographer} \emph{onstage} \textbf{songwriter song guitarist repertory}\\
& \textbf{voter vote votes election electoral polling poll presidential primaries} \emph{turnout}\\
& \textbf{film comedy} \emph{beginitalic} \emph{enditalic} \emph{sci} \textbf{filmmaker cinematic filmmaking movie starring}\\
\hline
\multirow{5}{*}{LDA-VAE}& \textbf{passenger destination traveler} \emph{fares} \emph{booking} \textbf{airlines luggage routes} \emph{rider} \emph{excursion}\\
& \textbf{acquisition shareholder merge} \emph{takeover} \emph{acquire} \textbf{merger} \emph{consolidated} \textbf{stockholder} \emph{suitor} \emph{consolidation}\\
& \textbf{soloist operatic composer} \emph{repertory} \textbf{troupe} \emph{choreographer} \emph{choreography} \textbf{sung} \emph{dances} \textbf{recital}\\
& \textbf{balloting nominating election elect incumbent victor primaries} \emph{contested} \textbf{electoral vote}\\
& \textbf{moviegoer studios filmmaker movies film filming} \emph{vh1} \textbf{studio stardom} \emph{rapper}\\
\hline
\multirow{5}{*}{NVDM}& \emph{nesting instructor ranchers wingspan veteran} \textbf{fly} \emph{manager} \textbf{pilot} \emph{ecosystems} \textbf{flight}\\
& \textbf{company billion companies} \emph{production} \emph{equipment} \emph{processed processing producer manufacturing products}\\
& \textbf{conducting conductor instrumental} \emph{interval staff} \textbf{discography} \emph{knighted radioactive charge} \textbf{director}\\
& \emph{degrees} \textbf{national party} \emph{billion} \textbf{nations} \emph{decrease university exceed disorder nuclear}\\
& \emph{bay} \textbf{film} \emph{indian french company} \textbf{novel} \emph{dec lake explorer travels}\\
\hline
\end{tabular}
}
\end{table*}

Typically topic models are evaluated based on the likelihood of held-out documents. However, as pointed out in \cite{chang2009reading}, higher likelihood of held-out document does not necessarily correspond to human judgement of topic coherence. In this subsection, we follow \cite{roder2015exploring} and choose five coherence metrics to evaluate the topics generated by models. They are C\_P (a metric based on a sliding window, a one-preceding segmentation of the given words and the confirmation measure of Fitelson's coherence), C\_A (a metric based on a context window, a pairwise comparison of the given words and an indirect confirmation measure that uses normalized pointwise mutual information and the cosine similarity), UCI (a metric based on a sliding window and the pointwise mutual information of all word pairs of the given topics), NPMI (an enhanced version of UCI using the normalized pointwise mutual information) and UMass \cite{mimno2011optimizing} (a metric based on document cooccurrence counts, a one-preceding segmentation and a logarithmic conditional probability as confirmation measure). For all these five metrics, higher value implies more coherent topic. In our evaluation, we choose the top 10 words to represent each topic and compute the topic coherence using the Palmetto library\footnote{https://github.com/dice-group/Palmetto}.

To compare the performance of the proposed approach, experiments are conducted on Grolier and NYtimes with five topic number settings [20, 30, 50, 75, 100]. The average coherence values are listed in Table \ref{tbs:average} and each value is computed by averaging the average topic coherences (all the topics are used) over five topic number settings. Besides, we calculate the average topic coherence among topics whose coherence values are ranked at the top 50\%, 70\%, 90\%, 100\% positions. For example, to calculate the average UCI coherence of ATM @ 70\%, we first compute the average UCI coherence with the select topics whose UCI values are ranked at the top 70\% positions for each topic number setting, and then average the five averaged coherence values. The corresponding results are shown in Figure \ref{fig:avg_bar}. It can be observed from Figure \ref{fig:avg_bar} that the proposed model outperforms the LDA, NVDM, LDA-VAE and ProdLDA in general.  {\color{black}This maybe caused by following factors: i) ATM models the multinomial distribution over topics using a Dirichlet prior, which is more proper than the Gaussian prior (used in NVDM) and logistic normal prior (used in LDA-VAE and ProdLDA). The usage of the Dirichlet prior in ATM make it could capture the multiplicity topic aspects from document collection and further obtain more coherent topics. 2.) The strong representation ability of the neural network makes the ATM could fit the true data distribution better than the traditional topic model and generate more coherent topics.}

\begin{table*}[!h]
\centering
\footnotesize
%\scriptsize
\label{table:stu}
\caption{The event examples extracted by ATM and LEM.}
\label{tbs:example_event}
\begin{tabular}{p{1.0cm}|c|l}
\hline
\multicolumn{1}{c|}{{\bfseries Events}}& \multicolumn{1}{c|}{\bfseries Method} &\multicolumn{1}{c}{\bfseries Representative Words}\\
\hline
\multirow{8}{*}{\makecell[c]{MH370}}&
\multirow{4}{*}{\makecell[cc]{ATM}}&org: \textbf{air airlines} ministry transport international \\
&&loc: \textbf{malaysia} beijing france vietnam dubai  \\
&&per: \textbf{hishammuddin hussein} najib kerry lee   \\
&&key: \textbf{search flight aircraft air plane}  \\
%\hline
\cline{2-3}
 &
\multirow{4}{*}{\makecell[tc]{LEM}}&org: \textbf{airlines air} international transport government     \\
&&loc: \textbf{malaysia} south korea beijing us    \\
&&per: \textbf{hussein hishammuddin} fitch long park  \\
&&key: \textbf{flight airlines plane} preliminary \textbf{search} \\
\hline

\multirow{8}{*}{\makecell[tc]{Saudi\\ MERS}}&
\multirow{4}{*}{\makecell[tc]{ ATM }}&org: \textbf{community} ministry \textbf{saudi} \textbf{healthcare} government   \\
&&loc: \textbf{saudi} ontario iran canada jeddah  \\
&&per: \textbf{president} obama jordan kerry walker   \\
&&key: \textbf{health hospital patients disease medical} \\
\cline{2-3} &
\multirow{4}{*}{\makecell[tc]{ LEM}}&org: \textbf{saudi} jordan army eastern state     \\
&&loc: east \textbf{saudi} jordan egypt israel \\
&&per: jordan \textbf{president} frank rob geldof    \\
&&key: east middle \textbf{respiratory syndrome health}    \\
\hline

\multirow{8}{*}{\makecell[tc]{Pakistan\\ vs. India}}&
\multirow{4}{*}{\makecell[tc]{ATM}} & org: \textbf{army kashmir} sharif taliban afghanistan    \\
&&loc: \textbf{pakistan kashmir india} afghanistan islamabad   \\
&&per: \textbf{sharif} kerry khan president lovell     \\
&&key: \textbf{army peace chief region} province  \\
\cline{2-3} &

\multirow{4}{*}{\makecell[tc]{LEM}}&org: \textbf{army kashmir} sharif government congress   \\
&&loc: \textbf{pakistan kashmir} islamabad \textbf{india} delhi  \\
&&per: \textbf{sharif} tsvangirai morgan dube biti   \\
&&key: \textbf{army chief} vein news \textbf{peace}    \\
\hline
\multirow{8}{*}{\makecell[tc]{Indian\\  Election}}&
\multirow{4}{*}{\makecell[tc]{ATM}}&org: \textbf{bjp party congress singh gandhi}     \\
&&loc: \textbf{gujarat india varanasi delhi} seemandhra \\
&&per: \textbf{modi singh} gandhi naidu khan    \\
&&key: \textbf{congress election candidate minister leader}   \\
\cline{2-3} &

\multirow{4}{*}{\makecell[tc]{ LEM}}&org: \textbf{bjp congress party commission delhi}  \\
&&loc: \textbf{delhi gujarat} modis \textbf{varanasi india}  \\
&&per: \textbf{modi} gandhi \textbf{singh} modis president  \\
&&key: \textbf{prime candidate election ministerial congress}  \\
\hline
\multirow{8}{*}{\makecell[tc]{Taksim\\  Clash}}&
\multirow{4}{*}{\makecell[tc]{ ATM}}&org: \textbf{police} city \textbf{government} erdogan \textbf{union}   \\
&&loc: \textbf{taksim istanbul} city \textbf{turkey union}     \\
&&per: \textbf{erdogan} park walker quinn hall     \\
&&key: \textbf{square protesters tear demonstrators street}      \\
\cline{2-3} &
\multirow{4}{*}{\makecell[tc]{ LEM}} & org: \textbf{police} international \textbf{labor} central greenpeace  \\
&&loc: \textbf{istanbul taksim turkey} rotterdam \textbf{union}   \\
&&per: mark \textbf{erdogan} geldof park hall      \\
&&key: \textbf{protesters square} international \textbf{gas water}  \\
\hline

\end{tabular}

\end{table*}

To explore how topic coherence results vary with different topic numbers, we show in Figure \ref{fig:avg_curve} the average topic coherence of two datasets vs. different topic number settings. We can observe that ATM achieves better results compared to other baselines most of the time with 20, 30, 50 or 75 topics. However, when the topic number is 100, the performance gap between ATM and LDA diminishes and in some cases (e.g., C\_P and C\_A for the Grolier dataset), ATM gives slightly worse results compared to LDA, though it still largely outperforms all the other baselines. This might attribute to the increased network complexity due to the larger topic number setting.

From the above topic coherence evaluation results, it is clear that ATM is able to extract more coherence topics compared to baselines. To verify this qualitatively, we show examples of topics from all the models in Table \ref{tbs:example_topics}. These topics correspond to \emph{`airline'}, \emph{`trade'}, \emph{`music'}, \emph{`election'} and \emph{`film'} respectively. Words that do not seem to belong to its corresponding topic are highlighted in italic. It can be observed that the number of less semantically relevant words somewhat correlates with the coherence results observed earlier in Table \ref{tbs:average} and Figure \ref{fig:avg_bar}.

Unlike traditional topic models, the proposed ATM could learn the semantic embeddings of words apart from generating coherent topics. The weights matrix $W_{w}\in \mathbb{R}^{V\times S}$ contains the word-level semantic information, and each row could be viewed as the corresponding word embedding. Thus, we select the topic words of six topics from a 50-topic run on the NYtimes corpus and use the Principal Component Analysis (PCA) to project their word embeddings into a two-dimensional space. The visualization of these topic words is shown as Figure \ref{fig:topic_visualization}. We can clearly see that the words related to the \emph{`trade'} topic are grouped at the lower right corner, and the topic words of \emph{`religious'} are displayed at the top region. Besides, the words related to the topics \emph{`music'} and \emph{`film'} are close to each other, which is not surprising, since these topics are closely related.

\subsection{Open Domain Event Extraction}

To further prove the feasibility of porting ATM to tasks other than topic modeling, we apply it for open domain event extraction. For this task, an event is represented in a structured form as $<org, loc, per, key>$\textsuperscript{\ref {olpk}} \cite{zhou2015unsupervised}, with each of the elements in the quadruples represented by a list of words.

We use the pre-identified named entities\textsuperscript{\ref{ner}}, verbs, nouns and adjectives to construct the word set of organization, location, person and keywords. When using ATM for event extraction, these four word sets and the event-specific word distribution are used to generate the related topics. For example, the organization topic of an event could be obtained by sorting the words in the organization word set based on the corresponding probabilities in the event-specific word distribution learned by ATM. Table \ref{tbs:example_event} shows the example events extracted by ATM and LEM where the relevant words are highlighted in bold. It can be observed that ATM performs comparably with LEM. However, while LEM required the model-specific inference algorithm to be derived, ATM did not need any modification of its network architecture or parameter estimation procedure.

To validate the correctness of the extracted events, we retrieve the title of articles using the event-related words from ATM and obtain the following results:
\begin{itemize}
\item \emph{Missing Malaysia Airlines flight MH370: Government report suggests official search for plane did not begin until four hours after disappearance.}
\item \emph{Saudi Arabia finds 26 more cases of MERS, Egypt reports first sufferer}.
\item \emph{India's defence experts and politicos condemn Pak Army Chief's Kashmir statement}.
\item \emph{Top BJP leaders, Rajnath Singh, MM Joshi, Sushma Swaraj to campaign for Narendra Modi in Varanasi}.
\item \emph{Turkey May Day protests hit by tear gas near Taksim Square - Panorama}. 
\end{itemize}

It is clear that the retrieved titles indeed correspond well with the extracted events by ATM. 

\section{Conclusions}
\label{sec:length}

We have proposed a novel topic modeling approach based on adversarial training. The proposed approach, ATM, models the topics with Dirichlet prior and employs the generator network to learn the semantic patterns among latent topics. Apart from automatically generating latent topics from a text corpus, it could also produce word-level semantic representations as a side product. The experimental comparison with the state-of-the-art methods show that ATM achieves improved topical coherence results. Moreover, the feasibility of porting ATM for tasks other than topic modeling has been verified for open domain event extraction. {\color{black}In the future, we want to incorporate the sequential information contained in texts into GAN based topic modeling approaches and devise a topic driven sentence generation model. And an extension to cope with the data sparsity in short text is also our future work. Besides,  another direction we are interested in exploring is to develop dynamic and correlated topic models based on adversarial training.}

\section{Acknowledgements}
We would like to thank anonymous reviewers for their valuable comments and helpful suggestions. {\color{black}This work was funded by the National Key Research and Development Program of China (2016YFC1306704), the National Natural Science Foundation of China (61772132), the Natural Science Foundation of Jiangsu Province of China (BK20161430).}

\bibliographystyle{unsrt}
\clearpage
\pagebreak
%\bibliography{nips}

\end{document}